\documentclass[runningheads]{llncs}
\usepackage{graphicx}
\usepackage{lscape}
\RequirePackage{rotating}
\usepackage{amsmath}
\usepackage{comment}
\usepackage{todonotes}
\usepackage{float}
\usepackage{makecell}
\usepackage{rotating} 
\usepackage{siunitx} 
\usepackage{booktabs}
\usepackage[bottom]{footmisc} 
\usepackage[breaklinks=true,hidelinks]{hyperref}

\usepackage{xstring}

\usepackage[utf8]{inputenc}

\begin{document}
\title{Putting RDF2vec in Order}

\author{Jan Portisch\inst{1,2}\orcidID{0000-0001-5420-0663} \and
Heiko Paulheim\inst{1}\orcidID{0000-0003-4386-8195}}
\authorrunning{J. Portisch and H. Paulheim}

\institute{Data and Web Science Group, University of Mannheim, Germany\\
\email{\{jan, heiko\}@informatik.uni-mannheim.de} 
\and
SAP SE Business Technology Platform | One Domain Model, Walldorf, Germany\\
\email{jan.portisch@sap.com}
}

\titlerunning{Putting RDF2vec in Order}

\maketitle
\begin{abstract}
The RDF2vec method for creating node embeddings on knowledge graphs is based on word2vec, which, in turn, is agnostic towards the position of context words. In this paper, we argue that this might be a shortcoming when training RDF2vec, and show that using a word2vec variant which respects order yields considerable performance gains especially on tasks where entities of different classes are involved.\footnote{Copyright © 2021 for this paper by its authors. Use permitted under Creative Commons License Attribution 4.0 International (CC BY 4.0).}
\vspace{0.3cm}
\\\textbf{Poster Submission}
\end{abstract}

\section{Introduction}

\begin{figure}[b]
    \centering
    \includegraphics[width=\textwidth]{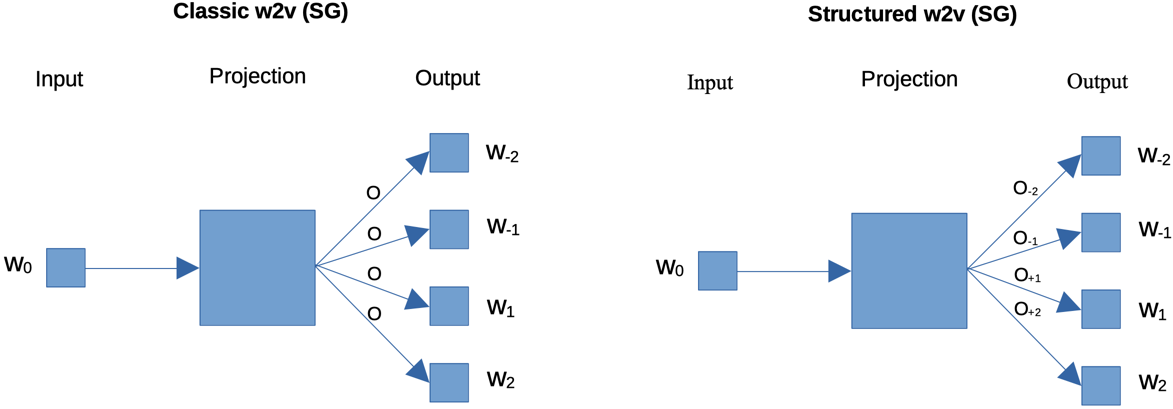}
    \caption{Classic word2vec vs. Structured word2vec}
    \label{fig:classic_vs_structured}
\end{figure}

\emph{RDF2vec}~\cite{DBLP:journals/semweb/RistoskiRNLP19} is a representation learning approach for entities in a knowledge graph. The basic idea is to first create \emph{sequences} from a knowledge graph by starting random walks from each node. These sequences are then fed into the \emph{word2vec} algorithm~\cite{mikolov2013distributed} for creating word embeddings, with each entity or property in the graph being treated as a ``word''. As a result, a fixed-size feature vector is obtained for each entity.

\emph{Word2vec} is a well-known neural language model to train latent representations (i.e., fixed size vectors) of words based on a text corpus.
Its objective is either to predict a word $w$ given its context words (known as continous bag of words  or CBOW), or vice versa (known as skip gram or SG).

Given the context $k$ of a word $w$, where $k$ is a set of preceding and succeeding words of $w$, the learning objective of word2vec is to predict $w$. This is known as \textit{continuous bag of words} model (CBOW). The \textit{skip-gram} (SG) model is trained the other way around: Given $w$, $k$ has to be predicted. Within this training process, the size of $k$ and is also known as \textit{window} or \textit{window size}. 

One shortfall of the original original word2vec approach is its insensitivity to the relative positions of words. It is, for instance, irrelevant whether a word is preceding or succeeding $w$, and the actual distance to $w$ is not considered. This property of word2vec is ideal to cope with the fact that in many languages, the same sentence can be expressed with different word orderings (cf. \emph{Yesterday morning, Tom ate bread} vs. \emph{Tom ate bread yesterday morning}). In contrast, walks extracted from knowledge graphs, the semantics of the underlying nodes differ depending on the position of an entity in the walk, as the following examples illustrates.

\begin{figure}[t]
    \centering
    \includegraphics[width=\textwidth]{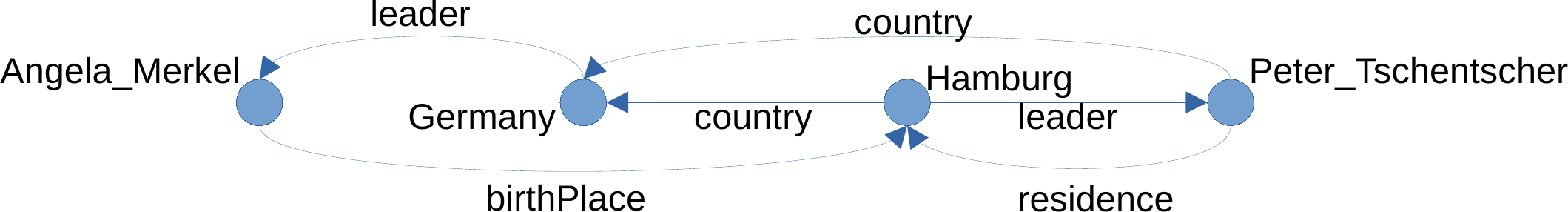}
    \caption{Example knowledge graph}
    \label{fig:example}
\end{figure}
Fig.~\ref{fig:example} depicts a small excerpt of a knowledge graph. Among others, the following walks could be extracted from the graph:

\footnotesize
\begin{verbatim}
Hamburg -> country -> Germany            -> leader     -> Angela_Merkel
Germany -> leader  -> Angela_Merkel      -> birthPlace -> Hamburg
Hamburg -> leader  -> Peter_Tschentscher -> residence  -> Hamburg
\end{verbatim}

\normalsize
If an RDF2vec model is trained for the entities in the center (i.e., \texttt{Germany}, \texttt{Angela\_Merkel}, and \texttt{Peter\_Tschentscher}), all of the sequences share exactly two entities in their context (\texttt{Hamburg} and \texttt{leader}), i.e., they will be projected equally close in the vector space. However, a model respecting positions would particularly differentiate the different meanings of \texttt{leader} (i.e., whether someone/thing \emph{has} or \emph{is} a leader), and the different \emph{roles} of involved entities (i.e., \texttt{Hamburg} as a place of birth or a residence of a person, or being located in a country). Therefore, it would map the two politicians closer to each other than to \texttt{Germany}. 

Ling et al.~\cite{DBLP:conf/naacl/LingDBT15} present an extension to the word2vec algorithm, known as \emph{structured word2vec}, which incorporates the positional information of words. This is achieved by using multiple encoders (CBOW) respectively decoders (SG) depending on the position of the context words. An illustration for SG can be found in Figure~\ref{fig:classic_vs_structured} where it is visible that the classic component uses only one output matrix $O$ which maps the embeddings to the output while the structured approach uses one output matrix per position in the window (e.g. $O_{+1}$ for the subsequent word to $w_0$).

In this paper, we present $RDF2vec_{oa}$, an \emph{order aware} variant of RDF2vec obtained by changing the training component from word2vec to structured word2vec, and show promising preliminary results.

\section{Related Work}
RDF2vec was one of the first approaches to adopt statistical language modeling techniques to knowledge graphs. Similar approaches, such as \emph{node2vec}~\cite{grover2016node2vec} and \emph{DeepWalk}~\cite{perozzi2014deepwalk}, were proposed for unlabeled graphs while knowledge graphs are labeled by nature, i.e., they contain different types of edges. 

Other language modeling techniques that have been adapted for knowledge graphs include GloVe~\cite{pennington2014glove}, which yielded \emph{KGlove}~\cite{DBLP:conf/semweb/CochezRPP17}, and BERT~\cite{devlin2018bert}, which yielded \emph{KG-BERT}~\cite{yao2019kg}. 

Variants of RDF2vec include the use of different heuristics for biasing the walks~\cite{DBLP:conf/wims/CochezRPP17}; \cite{DBLP:journals/corr/abs-2009-04404} evaluate multiple heuristics for biasing the walks or alternative walk strategies.
Very few authors tried to change the training objective of RDF2vec. 
Besides word2vec, the GloVe~\cite{DBLP:conf/emnlp/PenningtonSM14} algorithm has also been used~\cite{DBLP:conf/semweb/CochezRPP17}.

\section{Experiments and Preliminary Results}
We use jRDF2vec\footnote{\url{https://github.com/dwslab/jRDF2Vec}}~\cite{DBLP:conf/semweb/PortischHP20} to generate random walks and Ling et al.'s structured word2vec implementation\footnote{\url{https://github.com/wlin12/wang2vec}} to train an embedding based on the walks.

For the embeddings, we use the DBpedia 2016-04 dataset. We generated 500 random walks for each node in the graph with a depth of 4 (node hops).
word2vec and structured word2vec were trained using the same set of walks and the same training parameters: $SG$, $window=5$, and $size \in \{100, 200\}$.

We evaluate both, the classic and the position aware RDF2vec approach, on a variety of different tasks and datasets. For our evaluation, we use the \emph{GEval} framework~\cite{DBLP:conf/esws/PellegrinoAGRC20}. We follow the setup proposed in \cite{DBLP:conf/semweb/RistoskiVP16} and \cite{DBLP:conf/esws/PellegrinoAGRC20}. Those works use data mining tasks with an external ground truth. 
Different feature extraction methods -- which includes the generation of embedding vectors -- can then be compared using a fixed set of learning methods. 
Overall, we evaluate our new embedding approach on six tasks using 20 datasets altogether. The evaluation is conducted on six different downstream tasks -- classification and regression, clustering, determining semantic analogies, and computing entity relatedness and document similarity, the latter based on entities mentioned in the documents.

The results are presented in Table~\ref{tbl:results}. 
When comparing the classic to the order aware embeddings, it is visible that the performances are very similar on most tasks such as classification. 
A first observation is that we cannot observe significant performance drops on any of the tasks when switching from classic to order aware RDF2vec embeddings. 
However, significant performance increases can be observed on clustering tasks and on semantic analogy tasks, which are the tasks where entities of different classes are involved (whereas the classification and regression tasks deal with entities of the same class, e.g., cities or countries). The order aware RDF2vec configuration with 100 dimensions achieved on 7 datasets the overall best results and outperforms its classic configuration with the same dimension on 10 datasets partly with significantly better outcomes. On the other hand, in most cases where the classic variant performs better, it does so by a smaller margin. Thus, in general, the order-aware variant can be used safely without performance drops, and in some cases with significant performance gains.

\begin{table}[t]
\centering
\caption{Results of RDF2vec$_{classic}$ (c-100, c-200) and RDF2vec$_{oa}$ (oa-100, oa-200) trained with 100 and 200 dimensions respectively. The best value in each dimension group is printed in bold, the overall best value is additionally underlined.}
\label{tbl:results}
\footnotesize
\begin{tabular}{lll||cc|cc}
\toprule
Task & Metric & Dataset & c-100 & oa-100 & c-200 & oa-200\\
\toprule
Classification & ACC & AAUP              & \underline{\textbf{0.693}} & 0.679 & \textbf{0.692} & 0.683\\
               & ACC & Cities            & \textbf{0.793} & \textbf{0.793} & 0.798 & \underline{\textbf{0.807}}\\
               & ACC & Forbes            & \textbf{0.629} & 0.607 & \underline{\textbf{0.635}} & 0.630\\
               & ACC & Metacritic Albums & 0.783 & \underline{\textbf{0.799}} & 0.788 & \textbf{0.792}\\
               & ACC & Metacritic Movies & \textbf{0.757} & 0.736 & \underline{\textbf{0.763}} & 0.748\\
\hline
Clustering     & ACC & \makecell[l]{Cities/Countries (2k)} & 0.755 & \textbf{0.939} & 0.758 & \underline{\textbf{0.946}}\\
               & ACC & Cities/Countries      & \underline{\textbf{0.786}} & 0.785 & 0.7624 & \textbf{0.766}\\
               & ACC & \makecell[l]{Cities/Albums/Movies\\/AAUP/Forbes} & \underline{\textbf{0.932}} & 0.931 & 0.861 & \textbf{0.929}\\
               & ACC & Teams & 0.969 & \underline{\textbf{0.971}} & 0.892 & \textbf{0.945}\\
\hline
Regression     & RMSE & AAUP & 65.151 & \underline{\textbf{62.624}} & 66.301 & \textbf{65.077}\\
               & RMSE & Cities & 12.726 & \underline{\textbf{11.220}} & 14.855 & \textbf{13.484}\\
               & RMSE & Forbes & \underline{\textbf{34.290}} & 34.340 & 36.460 & \textbf{35.967}\\
               & RMSE & Metacritic Albums & 11.366 & \underline{\textbf{11.215}} & \textbf{11.528} & 11.651 \\
               & RMSE & Metacritic Movies & \textbf{19.091}  & 19.530 & \underline{\textbf{19.078}} & 19.432\\
\hline
Semantic & ACC & Capital-Countries & 0.852 & \underline{\textbf{0.990}} & 0.872 & \textbf{0.949} \\
Analogies                  & ACC & Capital-Countries (all) & 0.832 & \underline{\textbf{0.933}} & \textbf{0.901} & 0.896\\
                   & ACC & Currency-Country & 0.417 & \textbf{0.520} & \underline{\textbf{0.537}} & 0.441 \\
                   & ACC & City-State & 0.5577 & \textbf{0.607} & 0.555 & \underline{\textbf{0.627}}\\
\hline
\makecell[l]{Entity\\Relatedness} & \makecell[l]{Harmonic\\Mean} & - & \textbf{0.726} & 0.716 & \textbf{\textbf{0.747}} & \underline{\textbf{0.747}}\\
\hline
\makecell[l]{Document\\Similarity} & \makecell[l]{Kendall\\Tau} & - & \underline{\textbf{0.405}} & 0.373 & \textbf{0.350} & 0.325\\
\bottomrule
\end{tabular}
\end{table}

\section{Summary and Future Work}
\label{sec:summary_future_work}
In this paper, we presented a position aware variant of RDF2vec together with first very promising evaluation results. In the future, we plan to conduct more thorough analyses, analyzing which knowledge graph characteristics and downstream tasks benefit most from the ordered variant, and which do not. For example, we believe that graphs with a small set of predicates, or graphs which have all symmetric, inverse, and transitive relations materialized \cite{iana2020more}, can benefit more from using the ordered variant.

Furthermore, we plan to analyze how the ordered variant can be integrated into other RDF2vec configurations and flavours, such as different biased walks \cite{DBLP:conf/semweb/CochezRPP17}, or RDF2vec Light~\cite{DBLP:conf/semweb/PortischHP20}.

\bibliographystyle{splncs04}
\bibliography{references}

\end{document}